\documentclass[letterpaper]{article} 
\usepackage{aaai2026}  
\usepackage{times}  
\usepackage{helvet}  
\usepackage{courier}  
\usepackage[hyphens]{url}  
\usepackage{graphicx} 
\usepackage{natbib}  
\usepackage{caption} 
\usepackage{algorithm}
\usepackage{algorithmic}
\usepackage{booktabs}
\usepackage{multirow}
\usepackage{amssymb}
\usepackage{xcolor}
\usepackage{enumitem}
\usepackage{amsmath} 
\usepackage{booktabs} 
\usepackage{newfloat}
\usepackage{listings}
\usepackage{subfigure}
\DeclareCaptionStyle{ruled}{labelfont=normalfont,labelsep=colon,strut=off} 
\floatstyle{ruled}
\newfloat{listing}{tb}{lst}{}
\floatname{listing}{Listing}
\usepackage{color, xspace}

\title{Boosting Fine-Grained Urban Flow Inference via Lightweight Architecture \\and Focalized Optimization}

\author{
    Yuanshao Zhu\textsuperscript{\rm 1, 2},
    Xiangyu Zhao\textsuperscript{\rm 2, \footnotemark[1]},
    Zijian Zhang\textsuperscript{\rm 3},
    Xuetao Wei\textsuperscript{\rm 1},
    James Jianqiao Yu\textsuperscript{\rm 4,}
   \thanks{Corresponding authors}
}
\affiliations{
    \textsuperscript{\rm 1}Southern University of Science and Technology, China\\
    \textsuperscript{\rm 2}City University of Hong Kong, China\\
    \textsuperscript{\rm 3}Jilin University, China\\
    \textsuperscript{\rm 4}Harbin Institute of Technology, Shenzhen, China\\
    
    yuanshao@ieee.org, xianzhao@cityu.edu.hk, zhangzijian@jlu.edu.cn, weixt@sustech.edu.cn, jqyu@ieee.org
}

\usepackage{bibentry}

\begin{document}

\maketitle

\begin{abstract}
Fine-grained urban flow inference is crucial for urban planning and intelligent transportation systems, enabling precise traffic management and resource allocation.
However, the practical deployment of existing methods is hindered by two key challenges: the prohibitive computational cost of over-parameterized models and the suboptimal performance of conventional loss functions on the highly skewed distribution of urban flows.
To address these challenges, we propose a unified solution that synergizes architectural efficiency with adaptive optimization.
Specifically, we first introduce \textbf{PLGF}, a lightweight yet powerful architecture that employs a Progressive Local-Global Fusion strategy to effectively capture both fine-grained details and global contextual dependencies. 
Second, we propose \textbf{DualFocal} Loss, a novel function that integrates dual-space supervision with a difficulty-aware focusing mechanism, enabling the model to adaptively concentrate on hard-to-predict regions. 
Extensive experiments on 4 real-world scenarios validate the effectiveness and scalability of our method. 
Notably, while achieving state-of-the-art performance, PLGF reduces the model size by up to 97\% compared to current high-performing methods. Furthermore, under comparable parameter budgets, our model yields an accuracy improvement of over 10\% against strong baselines.
The implementation is included in the {https://github.com/Yasoz/PLGF}.
\end{abstract}



\section{Introduction}

Fine-grained urban flow data provides the foundational insights for modern intelligent transportation systems and smart city infrastructures, enabling applications like precise traffic management and responsive urban planning \cite{wang2020deep}.
However, acquiring such high spatial resolution data directly through dense sensor deployment is often infeasible due to prohibitive long-term costs for equipment, operation, and maintenance \cite{xie2020urban}. 
To address this, Fine-Grained Urban Flow Inference (FUFI), which infers fine-grained flow maps from available coarse-grained observations, has emerged as a pressing and cost-effective solution for developing sustainable and economically viable smart cities \cite{qu2022forecasting,wang2023fine}.

To accurately infer fine-grained urban flow, early FUFI methods draw from the super-resolution techniques of computer vision, where urban flow maps can be viewed as an image.
Foundational works like UrbanFM \cite{liang2019urbanfm} and UrbanPy \cite{ouyang2020fine} established the core paradigm, introducing deep learning frameworks with crucial spatial constraints and pyramid architectures to handle the mapping between coarse and fine-grained flows.
Follow-up efforts are mostly focused on refining architectural designs for better city-wide traffic profile capturing \cite{zhou2020enhancing}.
More recently, the field has shifted from pursuing pure accuracy to addressing practical deployment challenges, such as catastrophic forgetting in dynamic systems \cite{yu2023overcoming}, robustness to noisy data \cite{zheng2023diffuflow}, and cross-city knowledge transfer \cite{zheng2024fgitrans}.

\begin{figure}[!t]
    \centering
    \includegraphics[width=0.96\linewidth]{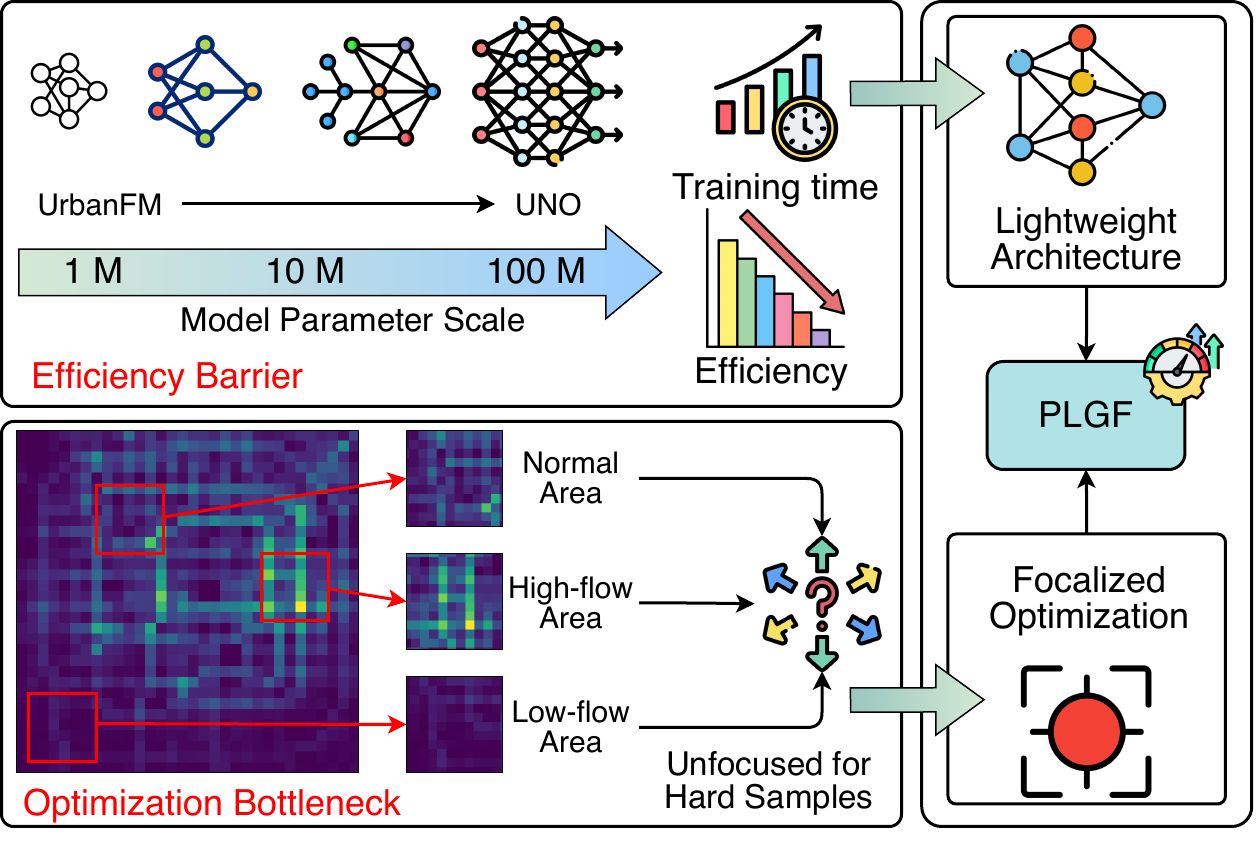} %
    \caption{The proposed PLGF addresses efficiency barriers and optimization bottleneck issues with a lightweight architecture design and focalized optimization.}
    \label{fig:intro}
    \vspace{-5mm}
\end{figure}

Despite these significant advancements, the very pursuit of higher accuracy has given rise to two critical limitations that hinder practical deployment, as shown in Figure \ref{fig:intro}.
First, an \textbf{efficiency barrier} has emerged. 
The push for higher performance has led to increasingly complex and over-parameterized architectures to capture intricate spatio-temporal dependencies \cite{gao2024enhancing}. 
The scale of model parameters has increased from 1 million to 100 million.
This trend towards ``model bloat'' incurs extreme computational costs for both training and inference, posing significant challenges for real-world applications.
Second, a persistent \textbf{optimization bottleneck} remains. 
Existing models are typically optimized with conventional regression losses, such as Mean Squared Error, which treat all prediction errors equally. 
By failing to account for the highly skewed, non-uniform nature of urban flow data, this generic approach leaves the optimization process unfocused, limiting the ultimate accuracy and robustness potential of the model, especially when dealing with high-variance samples \cite{liang2019urbanfm}.

To address these fundamental limitations, we propose a unified framework that rethinks both architectural design and optimization strategy.
First, we introduce \textbf{PLGF} (\textbf{P}rogressive \textbf{L}ocal-\textbf{G}lobal \textbf{F}usion), a lightweight yet powerful architecture designed for parameter efficiency. 
Specifically, PLGF adopts an efficient progressive local-global fusion framework and a context-aware integration mechanism.
This design enables it to capture complex multi-scale spatial dependencies with remarkable parameter efficiency, while ensuring that the entire process is subject to modulation by specific spatio-temporal contexts.
In addition, we propose the \textbf{DualFocal} Loss,  a novel loss function for focalized optimization. 
By integrating dual-scale (linear and logarithmic) supervision with a difficulty-aware focusing mechanism, our loss allows the model to adaptively concentrate on hard-to-predict, low-flow regions and high-variance samples that are often ignored by traditional methods.

The main contributions of this paper are as follows:
\begin{itemize}[leftmargin=*]
    \item We propose PLGF, a novel, lightweight, and parameter-efficient architecture. It effectively captures multi-scale spatial dependencies based on a progressive local-global fusion and context-aware strategy, thus significantly reducing the computational cost.

    \item We introduce DualFocal Loss, a universal and plug-and-play loss function with a difficulty-aware focusing mechanism. It can be flexibly applied to various FUFI models, improving both accuracy and robustness.

    \item Extensive experiments on 4 real-world scenarios demonstrate that 
    PLGF achieves over 10\% performance improvement under the same parameter budget and reduces parameter count by approximately 97\% while maintaining state-of-the-art performance.

\end{itemize}

\section{Related Work}
Fine-grained urban flow inference seeks to recover high-resolution urban flow maps from coarse-grained observations.
The foundational paradigm was established by UrbanFM \cite{liang2019urbanfm}, which introduced a distributional upsampling module enforcing strict spatial constraints, ensuring that the sum of fine-grained flows matches their corresponding coarse-grained region.
Subsequent works advanced FUFI from several perspectives. 
UrbanPy \cite{ouyang2020fine} proposed a cascading pyramid architecture for progressive upsampling, improving scalability to large-scale urban maps. 
DeepLGR \cite{liang2021revisiting} further enhanced spatial modeling by introducing a dual-path network that jointly learns global and local features. 
Alternative approaches, such as FODE \cite{zhou2020enhancing} and UrbanODE \cite{zhou2021inferring}, leveraged neural ordinary differential equations to capture continuous urban dynamics and address numerical instability.
More recently, research has shifted to practical deployment challenges in dynamic, real-world environments. 
CUFAR \cite{yu2023overcoming} addressed catastrophic forgetting in dynamic environments via an adaptive knowledge replay strategy, while UNO \cite{gao2024enhancing} proposed data-free incremental learning using neural operators for privacy-preserving, scale-invariant modeling. 
Robustness to data quality has also been explored: multi-task frameworks have been developed for simultaneous missing data imputation and inference \cite{li2022fine}, and denoising strategies have been integrated to enhance resilience to noisy sensor inputs \cite{zheng2023diffuflow}. 
To tackle data scarcity in target cities, cross-city transfer learning methods have been introduced to leverage knowledge from data-rich source cities \cite{zheng2024adatm, zheng2024fgitrans}.

Despite these advancements, most existing approaches remain limited by large model sizes and high computational costs. 
Furthermore, optimization strategies tailored to the highly skewed and non-uniform nature of urban flow data are still lacking. 
In this work, we address these gaps by proposing a lightweight architecture and a focused optimization strategy, aiming to improve both efficiency and adaptability for real-world FUFI applications.

\section{Preliminary}

This section formally defines the key concepts and the problem of FUFI, providing the foundation for our methodology.

\noindent \textbf{Urban Flow Map.}
Following standard practice, a city or a region of interest is partitioned into a uniform $I \times J$ grid map based on longitude and latitude.
At a given time interval, the traffic flow across the whole area is represented by an urban flow map, denoted as $X \in \mathbb{R}^{I \times J}$, where each element $x_{i,j} \in \mathbb{R}_{+}$ indicates the total flow volume (e.g., vehicles, pedestrians) within the corresponding grid cell $(i, j)$.

\noindent \textbf{Coarse-grained and Fine-grained Urban Flow.}
Urban flow maps can be constructed at varying spatial granularities depending on the scale of observation.
A fine-grained flow map $X_f \in \mathbb{R}^{(N_f \times N_f)}$ provides detailed mobility information, while a coarse-grained flow map $X_c \in \mathbb{R}^{(N_c \times N_c)}$ offers an aggregated perspective. 
Each cell in the coarse-grained map corresponds to a non-overlapping $N \times N$ block of fine-grained cells, where $N \in \mathbb{Z}{+}$ is the upscaling factor.
This hierarchical aggregation imposes a crucial spatial constraint: the total flow in any coarse cell equals the sum of flows in its constituent fine-grained cells. Formally:
\begin{equation}
      x^c_{i,j} = \sum_{i^{\prime}=N_i}^{N_{(i+1)}-1} \sum_{j^{\prime}=N_j}^{N_{(j+1)}-1} x^{f}_{i^{\prime},j^{\prime}},
\end{equation}
where $x^c_{i,j}$ is the flow in coarse cell $(i, j)$, and $x^f_{i',j'}$ are the flows in the corresponding fine-grained cells.

\noindent \textbf{Fine-Grained Urban Flow Inference.}
Given an observed coarse-grained urban flow map $X_c \in \mathbb{R}^{(I \times J)}$ and an integer upscaling factor $N$, the goal of FUFI task is to learn an inference function $F_{\theta}$ that can accurately reconstruct the corresponding high-resolution, fine-grained flow map $X_f \in \mathbb{R}^{(NI \times NJ)}$. 
The predicted map $\hat{X}_f = F_{\theta}(X_c)$ should closely approximate the true fine-grained map while strictly satisfying the spatial constraint above.
Optionally, the inference can incorporate external factors $E$ (such as time of day, weather), yielding the formulation $\hat{X}_f = F_{\theta}(X_c, E)$.

\section{Methodology}
\begin{figure}[!t]
    \centering
    \includegraphics[width=0.96\linewidth]{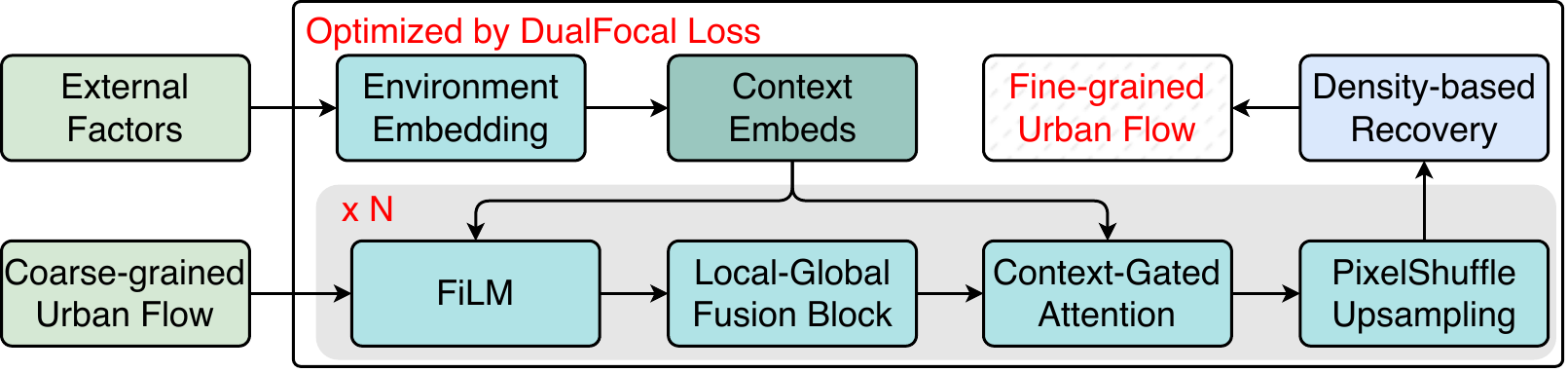} %
    \caption{The pipeline of PLGF architecture.}
    \label{fig:pipeline}
    \vspace{-3mm}
\end{figure}

As illustrated in Figure~\ref{fig:pipeline}, our proposed PLGF framework is a conditional deep network designed to infer fine-grained urban flow maps from coarse-grained inputs and external factors.
The entire design is driven by two principles carefully chosen to address the core challenges of efficiency and optimization: a lightweight architecture built on progressive learning and a strategy for adaptive optimization via a novel loss function.
The progressive learning principle tackles the challenge of high-ratio $N$-times upsampling by decomposing it into $\log_2(N)$ sequential $2$-times upsampling stages.
This approach is inherently more parameter-efficient and stable than a single-step upsampling process. At each stage, context-aware modulation ensures the model dynamically adapts to external conditions. 
The entire framework is optimized end-to-end with our proposed DualFocal Loss, which forces the model to adaptively focus on the most challenging aspects of the urban flow data distribution.

\subsection{PLGF: A Lightweight Fusion Architecture}
The PLGF architecture materializes our design principles into a three-part pipeline as shown in Figure \ref{fig:framework}: (1) an \textit{Environment Context Embedding} block to process external factors, (2) a series of stacked \textit{Progressive Upscaling Blocks (PUBs)} to iteratively refine features and increase resolution, and (3) a final \textit{Density-based Recovery block} to ensure spatial constraint, fine-grained output.

\subsubsection{Environment Context Embedding.}
Firstly, urban flow is not static, which is heavily influenced by a myriad of external factors such as time, weather, and public events. A robust FUFI model must be able to adapt its predictions to these changing conditions. To achieve this, we need a mechanism to encode these heterogeneous external factors into a unified and powerful conditioning signal.

The Environment Context Embedding module transforms raw external factors $E \in \mathbb{R}^7$ into a unified, high-dimensional conditional vector $\boldsymbol{e}_{\text{cond}}$.
The categorical features (e.g., day of the week, hour) are mapped to dense embeddings $\{\boldsymbol{v}_{\text{day}}, \boldsymbol{v}_{\text{hour}}, \boldsymbol{v}_{\text{weather}}\}$, while continuous features (e.g., temperature) are processed by a multi-layer perceptron (MLP) to form $\boldsymbol{v}_{\text{cont}}$. 
To capture interdependencies adaptively, these embeddings are passed through a multi-head self-attention (MHA) layer. The output is then aggregated and projected to produce the final conditional vector:
\begin{equation}
    \boldsymbol{e}_{\text{cond}} = \text{Linear}(\text{MHA} ([ \boldsymbol{v}_{\text{day}}, \boldsymbol{v}_{\text{hour}}, \boldsymbol{v}_{\text{weather}}, \boldsymbol{v}_{\text{cont}} ])),
\end{equation}
which $\boldsymbol{e}_{\text{cond}} \in \mathbb{R}^{d}$ serves as a dynamic conditioning signal throughout the subsequent progressive upsampling stages.

\subsubsection{Progressive Upscaling Block.}
\begin{figure*}[t]
    \centering
    \includegraphics[width=0.96\linewidth]{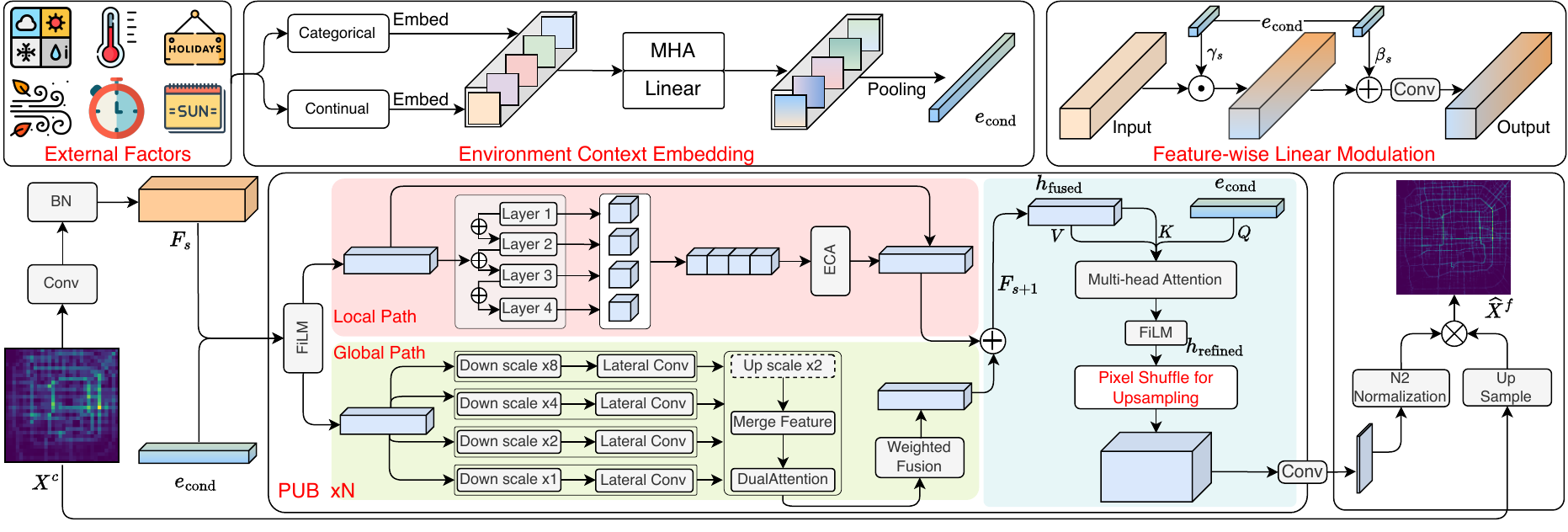} %
    \caption{The overall details of PLGF architecture, which consists of four main components, extract spatio-temporal features using a progressive and context-aware manner.}
    \label{fig:framework}
    \vspace{-3mm}
\end{figure*}

As illustrated in Figure \ref{fig:framework}, the core of the PLGF framework is the Progressive Upscaling Block (PUB).
Given initial features $\boldsymbol{F}_0$ extracted from the input convolutional layer processing the coarse-grained flow $X_c$, the PUB iteratively refines and upsamples these features, dynamically conditioned on the context vector $\boldsymbol{e}_{\text{cond}}$:
\begin{equation}
\boldsymbol{F}_s = \text{PUB}_s(\boldsymbol{F}_{s-1},~\boldsymbol{e}_{\text{cond}}), \quad s = 1, \ldots, S,
\end{equation}
where $\boldsymbol{F}_s \in \mathbb{R}^{C \times (2^s H) \times (2^s W)}$, and $C, H, W$ denote the channel, height, and width of the initial feature map, respectively.
Each PUB is composed of four main components: a Feature-wise Linear Modulation (FiLM) for context-aware feature adaptation, a Local-Global Fusion Block for comprehensive feature extraction, a Context-Gated Attention Block for feature refinement, and finally a PixelShuffle Block for  $2\times$ spatial upsampling.

\noindent \textbf{FiLM for Context-Aware Modulation.}
At the start of each PUB, a FiLM layer \cite{perez2018film} injects contextual information into the feature maps.
Given the feature map $\boldsymbol{F}_s$ from the previous stage and the conditional vector $\boldsymbol{e}_{\text{cond}}$, the FiLM layer generates channel-wise scaling $\gamma_s$ and shifting parameters $\beta_s$ via linear projections:
\begin{align}
\boldsymbol{F}_{s-1}^{\prime} &= (1 + \gamma_s) \odot \boldsymbol{F}_{s-1} + \beta_s\\
\gamma_s = &f_s(\boldsymbol{e}_{\text{cond}}),~~~\beta_s = h_s(\boldsymbol{e}_{\text{cond}}), 
\end{align}
where $\odot$ denotes element-wise multiplication.
This dynamic modulation enables the network to adapt feature responses at each stage to the specific environmental context.

\noindent \textbf{Local-Global Fusion Block.}
Urban flow patterns exhibit complex spatial dependencies. Predicting the flow in one grid cell requires understanding both its immediate surroundings (e.g., local traffic) and its role within the larger city-wide traffic network (e.g., its position). Capturing these multi-scale spatial features simultaneously is a major challenge.
Therefore, we design a dual-path block to explicitly capture both local and global features.

\noindent \textit{\textbf{Local Path: Efficiently Capturing Fine-Grained Details}.}
The local path is engineered to capture detailed, hierarchical features within local neighborhoods. At its core is a carefully optimized Residual Dense Block (RDB), a choice motivated by its high parameter efficiency.
The architecture of our RDB is motivated by several key principles aimed at maximizing both performance and efficiency.
The efficiency of the RDB stems from its use of dense connectivity. As shown in Figure \ref{fig:framework}, the output of each convolutional layer is concatenated with the inputs of all subsequent layers within the block.
This mechanism encourages feature reuse, allowing the network to build highly discriminative local representations without needing to learn redundant feature maps, which is key to its lightweight nature.
For an input $\boldsymbol{x}_0$, the operation of the $l$-th layer $H_l$ is:
\begin{align}
\boldsymbol{x}_l = H_l([\boldsymbol{x}_0 \oplus \boldsymbol{x}_1 \oplus \dots \oplus \boldsymbol{x}_{l-1}]),
\end{align}
where $[\oplus ]$ denotes the concatenation. 
To enhance training stability and performance, we adopt Group Normalization and the GELU activation function.

After the dense feature extraction, the collection of generated feature maps is fused via a $1\times1$ convolution. To further enhance discriminative power, we then apply an Efficient Channel Attention (ECA) layer \cite{wang2020eca}.
This lightweight attention mechanism adaptively recalibrates channel-wise feature responses, amplifying informative features while suppressing less useful ones. 
Finally, a residual connection adds the refined features back to the original input $\boldsymbol{x}_0$. 
This strategy stabilizes training and focuses the block on learning only the essential residual information. The entire RDB operation is summarized as:
\begin{equation}
\text{RDB}(\boldsymbol{x}_0) = \boldsymbol{x}_0 + \alpha \cdot \text{ECA}(f_{\text{fusion}}([\boldsymbol{x}_0 \oplus \boldsymbol{x}_1 \oplus \dots \oplus \boldsymbol{x}_{l}])),
\end{equation}
where $\alpha$ is a small residual scaling factor (initialized with 0.1) to stabilize the training process.

\noindent \textit{\textbf{Global Path: Adaptively Aggregating Multi-Scale Context}.}
The global path is designed to capture long-range dependencies by aggregating context from multiple spatial scales. We employ an Enhanced Feature Pyramid Network (EnhancedFPN), which augments the standard FPN \cite{Lin2017fpn} with adaptive attention mechanisms.
Given the input feature map $\boldsymbol{x}_{0}$, EnhancedFPN first constructs a feature pyramid  $\{\boldsymbol{p}_{0}, \boldsymbol{p}_{1}, \dots, \boldsymbol{p}_{s-1}\}$ using adaptive average pooling at progressively larger scales $\boldsymbol{p}_i = \textbf{AvgPool}(\boldsymbol{x}_0, 2^i)$.
This provides multiple views of the input at different levels of granularity. Following this, a top-down fusion pathway systematically enriches finer-grained feature maps (e.g., $\boldsymbol{p}_{i}$) by upsampling and merging them with semantic context from coarser feature maps (e.g., $\boldsymbol{p}_{i+1}$).
This process systematically enriches the finer-grained features with a higher level of semantic context than the coarser-grained ones, creating a semantically rich feature hierarchy.

Crucially, to make this fusion process adaptive, a Dual Attention module is applied at each fusion level. It uses channel and spatial attention to refine salient flow patterns and spatial structures, focusing on the most relevant information at each scale $\{\boldsymbol{p}^{\prime}_{0}, \boldsymbol{p}^{\prime}_{1}, \dots, \boldsymbol{p}^{\prime}_{s-1}\}$. 
After refining all levels, the final output is produced via an adaptive weighted fusion of all pyramid feature maps:
\begin{equation}
\text{EnhancedFPN}(\mathbf{x}_0) = \sum_{i=0}^{s-1} \sigma(\beta_i) \cdot  \boldsymbol{p}^{\prime}_{i},
\end{equation}
where $\sigma(\cdot)$ is the sigmoid function and the learnable weights ${\beta_i}$ allow the network to adaptively emphasize the most informative scales for a given input. 
The outputs from both local and global paths are concatenated and fused to integrate fine-grained details with multi-scale global context.

\noindent \textbf{Context-Gated Attention.}
After fusing local and global features, a final, precise refinement is needed. Traditional methods often use separate, heavy convolutional layers for each time period to integrate external information \cite{liang2019urbanfm, gao2024enhancing}. 
This reduces computational efficiency and significantly increases model parameters.
We employ a Context-Gated Attention block for this final, adaptive refinement. 
This block uses the powerful mechanism of multi-head attention to dynamically gate the fused features   $\boldsymbol{h}_{\text{fused}}$ based on the context vector $\boldsymbol{e}_{\text{cond}}$.
Specifically, the context vector serves as the ``query'', while the feature map acts as the ``key'' and ``value'':
\begin{align}
    \boldsymbol{v}_{\text{agg}} = \text{MHA}(\boldsymbol{e}_{\text{cond}}, \boldsymbol{h}_{\text{fused}}, \boldsymbol{h}_{\text{fused}}).
\end{align}
The resulting vector, $\boldsymbol{v}_{\text{agg}}$, represents a contextually-weighted aggregation of the most salient spatial information. 
We then use this aggregated vector to generate a new set of FiLM-style parameters ($\gamma$, $\beta$).
This performs a second, deeper modulation on the original fused features: $\boldsymbol{h}_{\text{refined}} = \gamma \odot \boldsymbol{h}_{\text{fused}} + \beta$.
This two-step process ensures that the features are not just broadly conditioned, but are precisely and adaptively refined based on a deep, contextual understanding of the fused local-global information.

\noindent \textbf{Spatial Upsampling with PixelShuffle.}
The final component of the PUB is responsible for upsampling the spatial resolution by a factor of $2$.
We adopt the PixelShuffle layer \cite{shi2016pixel}, as it is computationally efficient and effectively mitigates checkerboard artifacts.
Specifically, the PixelShuffle operation first increases the number of feature channels through a standard convolution, then rearranges channel elements into the spatial dimensions, effectively doubling both height and width:
\begin{equation}
\boldsymbol{F}_s = \text{PixelShuffle}(\boldsymbol{h}_{\text{refined}}).
\end{equation}
The resulting feature map $\boldsymbol{F}_s$ serves as the input for the next progressive stage, or, in the final stage, is passed to the output generation module.

\subsubsection{Density-based Recovery.}
A key physical constraint in FUFI is that the sum of flows in the fine-grained sub-regions must equal the observed flow in the corresponding coarse-grained region. Any valid model must strictly adhere to this conservation law.
Following the final PUB, a density-based recovery module ensures this constraint is met. It first uses a convolutional layer with a ReLU activation to produce a non-negative raw density map. Then, the parameter-free N2-Normalization layer \cite{liang2019urbanfm} normalizes the densities within each super-region to sum to one, creating a valid probability distribution. 
The final fine-grained flow map $\hat{X}_f$ is obtained by element-wise multiplication of this distribution with the upsampled coarse-grained flow $X_c$.
This process guarantees both physical consistency and fine-grained expressiveness.
\begin{equation}
\hat{X}_f = \text{Norm}(\text{ReLU}(\text{Conv}(\boldsymbol{F}_s))) \odot \text{Upsample}(X_c).
\end{equation}

\subsection{Focalized Optimization Strategy}
A fundamental challenge in urban flow modeling is the extreme data imbalance. A few central areas exhibit extremely high flow, while the vast majority of regions have very low or near-zero flow. Standard regression losses like Mean Squared Error (MSE) are dominated by the large absolute errors in high-flow regions. 
This biases the model towards fitting these few areas, while neglecting the subtle but critical patterns in the more prevalent low-flow regions. Consequently, the model's performance on relative metrics (like MAPE) and its overall robustness suffer.
To address this, we propose \textbf{ DualFocal Loss}, a novel loss function tailored for such skewed regression tasks. It integrates two key ideas: dual-scale supervision and hard sample mining.

\noindent \textbf{Dual-Scale Supervision.}
To ensure the model remains sensitive to errors across the entire spectrum of flow magnitudes, we integrate supervision at two complementary scales: the linear scale and the logarithmic scale.
(1) \textit{Linear scale}:  We compute a standard L1 loss between the predicted flow $\hat{X}_f$ and the target ${X}_f$,i.e., $\mathcal{L}_{\text{l1}} = |\hat{X}_f - X_f|$.
This term emphasizes accuracy in absolute flow values, particularly benefiting high-flow regions.
(2) \textit{Logarithmic scale}: To enhance sensitivity in low-flow regions, we introduce a second component, $\mathcal{L}_{\text{log}} = |\log(1+\hat{X}_f) - \log(1+X_f)|$, which applies the L1 norm after a $\log(1+x)$ (ensures numerical stability for zero flows) transformation.  
The logarithmic compression of the large dynamic range of urban flow amplifies the relative importance of errors in low-flow areas. This encourages the model to achieve high fidelity in both bustling centers and quiet residential zones.
The combined dual-scale loss is formulated as a weighted sum $\mathcal{L}_{\text{ds}} = \mathcal{L}_{\text{l1}} + \lambda \cdot \mathcal{L}_{\text{log}}$,
where $\lambda$ balances the contribution of the two scales.

\noindent \textbf{Focalized for Hard Sample Mining.}
While dual-scale supervision balances the learning across regions with varying flow magnitudes, it treats all predictions within those regions equally.
To compel the model to focus on more difficult predictions, we introduce a focal mechanism inspired by Focal Loss \cite{lin2017focal}. We dynamically re-weight the contribution of each sample to the loss based on its current prediction difficulty (i.e., its $\mathcal{L}_{\text{ds}}$ value).
\begin{equation}
\mathcal{L}_{\text{Dual-Focal}} = (f(\beta \cdot \mathcal{L}_{\text{ds}}))^\gamma \cdot \mathcal{L}_{\text{ds}},
\end{equation}
where the term $(f(\beta \cdot \mathcal{L}_{\text{ds}}))^\gamma$ is a modulating factor that increases the weight of hard-to-predict samples (those with large $\mathcal{L}_{\text{ds}}$). 
This forces the model to allocate more of its capacity to overcome challenging cases, improving overall accuracy and robustness.
By jointly leveraging dual-scale supervision and focalized re-weighting, the proposed Dual-Focal Loss provides a comprehensive optimization strategy tailored to the unique challenges of FUFI task.

\section{Experiments}

\begin{table*}[!t]
\centering
\small
\label{tab:my_label}
\begin{tabular}{lcccccccccccc}
\toprule
& \multicolumn{3}{c}{\textbf{Task-1}} & \multicolumn{3}{c}{\textbf{Task-2}} & \multicolumn{3}{c}{\textbf{Task-3}} & \multicolumn{3}{c}{\textbf{Task-4}}  \\
\cmidrule(lr){2-4} \cmidrule(lr){5-7} \cmidrule(lr){8-10} \cmidrule(lr){11-13}
& MSE & MAE & MAPE & MSE & MAE & MAPE & MSE & MAE & MAPE & MSE & MAE & MAPE \\
\midrule
SRCNN & 18.464 & 2.491 & 0.714 & 21.270 & 2.681 & 0.689 & 23.184 & 2.829 & 0.727 & 14.730 & 2.289 & 0.665 \\
ESPCN & 17.690 & 2.497 & 0.732 & 20.875 & 2.727 & 0.732 & 22.505 & 2.862 & 0.773 & 13.898 & 2.228 & 0.711 \\
VDSR & 17.297 & 2.213 & 0.467 & 21.031 & 2.498 & 0.486 & 22.372 & 2.548 & 0.461 & 13.351 & 1.978 & 0.411 \\
DeepSD & 17.272 & 2.368 & 0.614 & 20.738 & 2.612 & 0.621 & 22.014 & 2.739 & 0.682 & 15.031 & 2.297 & 0.652 \\
SRResNet & 17.338 & 2.457 & 0.713 & 20.466 & 2.660 & 0.688 & 21.996 & 2.775 & 0.717 & 13.446 & 2.189 & 0.637 \\
\midrule
UrbanFM & 16.372 & 2.066 & 0.335 & 19.548 & 2.284 & 0.328 & 21.243 & 2.398 & 0.336 & 12.744 & 1.850 & 0.311   \\
DeepLGR & 17.125 & 2.103 & 0.339 & 21.217 & 2.386 & 0.350 & 23.563 & 2.497 & 0.351 & 13.390 & 1.916 & 0.345   \\
FODE & 16.473 & 2.142 & 0.403 & 19.884 & 2.377 & 0.395 & 21.425 & 2.490 & 0.417 & 12.840 & 1.947 & 0.396  \\
UrbanODE & 16.342 & 2.135 & 0.406 & 19.648 & 2.357 & 0.394 & 21.177 & 2.460 & 0.408 & 12.668 & 1.929 & 0.391   \\
UrbanPy & 16.082 & 2.026 & 0.329 & 19.025 & 2.232 & 0.318 & 20.810 & 2.333 & 0.313 & 12.336 & 1.810 & 0.304  \\
CUFAR & 14.991 & 1.952 & 0.306 & 18.259 & 2.186 & 0.301 & 19.309 & 2.243 & 0.289 & 11.681 & 1.758 & 0.288  \\
UNO & 14.691 & 1.927 & 0.297 & 17.722 & 2.148 & 0.290 & 19.072 & 2.217 & 0.279 & 11.514 & 1.736 & 0.276 \\
\midrule   
PLGF & \textbf{14.408} & \textbf{1.890} & \textbf{0.281} & \textbf{17.364} & \textbf{2.098}  & \textbf{0.274} & \textbf{18.765} & \textbf{2.179}  & \textbf{0.268} &  \textbf{11.327} & \textbf{1.716} & \textbf{0.270}  \\
\midrule
\textbf{$\Delta$ (\%)} &  
$\downarrow$1.93 & $\downarrow$1.92 & $\downarrow$5.39 & $\downarrow$2.02 & $\downarrow$2.33 & $\downarrow$5.52 & $\downarrow$1.61 & $\downarrow$1.71 & $\downarrow$3.94 & $\downarrow$1.62 & $\downarrow$1.15 & $\downarrow$2.17\\
\bottomrule
\end{tabular}
\caption{Overall performance comparison with baseline methods.}
\label{tab:overall}
\end{table*}

In this section, we conduct extensive experiments to validate our proposed framework. We first benchmark PLGF against state-of-the-art methods to demonstrate its superiority.
Subsequently, we present extensive ablation studies to dissect the contribution of each key architectural and optimization component, and analyze the model's overall efficiency.

\subsection{Experimental Setup}

\noindent \textbf{Datasets.}
To comprehensively assess the effectiveness of our proposed framework, we conduct experiments on the widely used public TaxiBJ benchmark dataset \cite{liang2019urbanfm}. 
This large-scale dataset records city-wide taxi flows in Beijing over four distinct periods, each often considered as an independent real-world scenario, we denote as \textbf{Task-1} to \textbf{Task-4}.
Each data sample consists of a coarse-grained flow map ($32 \times 32$) and its corresponding fine-grained ground truth of ($128 \times 128$), and an external feature vector such as weather, time, and other contextual information.
We follow the standard data partitioning protocol: the temporally ordered data is split into training, validation, and test sets with a ratio of 7:1:2, ensuring reproducibility.

\noindent \textbf{Baselines.}
We benchmark our model against a broad range of representative methods.
First, we include several classic methods from the computer vision domain to establish a foundational baseline, including SRCNN \cite{dong2015image}, ESPCN \cite{shi2016pixel}, VDSR \cite{kim2016accurate}, DeepSD \cite{vandal2017deepsd}, and SRResNet \cite{ledig2017photo}.
Also, we compare against a set of SOTA models designed specifically for the FUFI task, including UrbanFM \cite{liang2019urbanfm}, UrbanPy \cite{ouyang2020fine}, DeepLGR \cite{liang2021revisiting}, FODE \cite{zhou2020enhancing}, UrbanODE \cite{zhou2021inferring}, CUFAR \cite{yu2023overcoming}, and UNO \cite{gao2024enhancing}.


\noindent \textbf{Metrics.} 
In line with standard practice, we evaluate model performance using three widely adopted metrics, which together capture both absolute and relative prediction accuracy:
Mean Square Error (MSE), Mean Absolute Error (MAE), and Mean Absolute Percentage Error (MAPE), where lower values indicate better performance. 

\noindent \textbf{Implementation.} 
All experiments are conducted on a server equipped with NVIDIA 4090 GPUs, and the results are reported as the average of five runs.

\subsection{Overall Performance}
Table~\ref{tab:overall} reports the comprehensive performance comparison of our proposed PLGF framework against a suite of baselines.
The results clearly demonstrate the superiority of our approach. As expected, FUFI-specific models consistently outperform general-purpose super-resolution methods, underscoring the necessity of domain-specific architectural designs. Among the FUFI-specific baselines, recent competitive models such as CUFAR and UNO demonstrate the most competitive performance, setting a high benchmark for accuracy.
Notably, compared to the best prior baseline (UNO), PLGF yields relative improvements of 1.6\%--2.0\% in MSE, 1.2\%--2.3\% in MAE, and up to 2.1\%--5.5\% in MAPE across different tasks. This pronounced reduction in MAPE is particularly significant, as it demonstrates PLGF’s superior ability to capture subtle, low-activity flow patterns, directly addressing the core challenge of the highly imbalanced, non-uniform data distributions motivating our work. 
Furthermore, the consistent improvements across all temporal splits highlight the robustness and generalization capability of our approach, confirming that the architectural innovations and the Focalized optimization strategy not only enhance overall accuracy but also ensure reliable performance in both high-flow and low-flow regions. 
Collectively, by explicitly addressing the imbalance and context sensitivity of urban flow data, PLGF achieves new SOTA and sets a strong foundation for future urban computing research.

\subsection{Efficiency Evaluation}

\begin{figure}[h]
    \centering
    \includegraphics[width=0.9\linewidth]{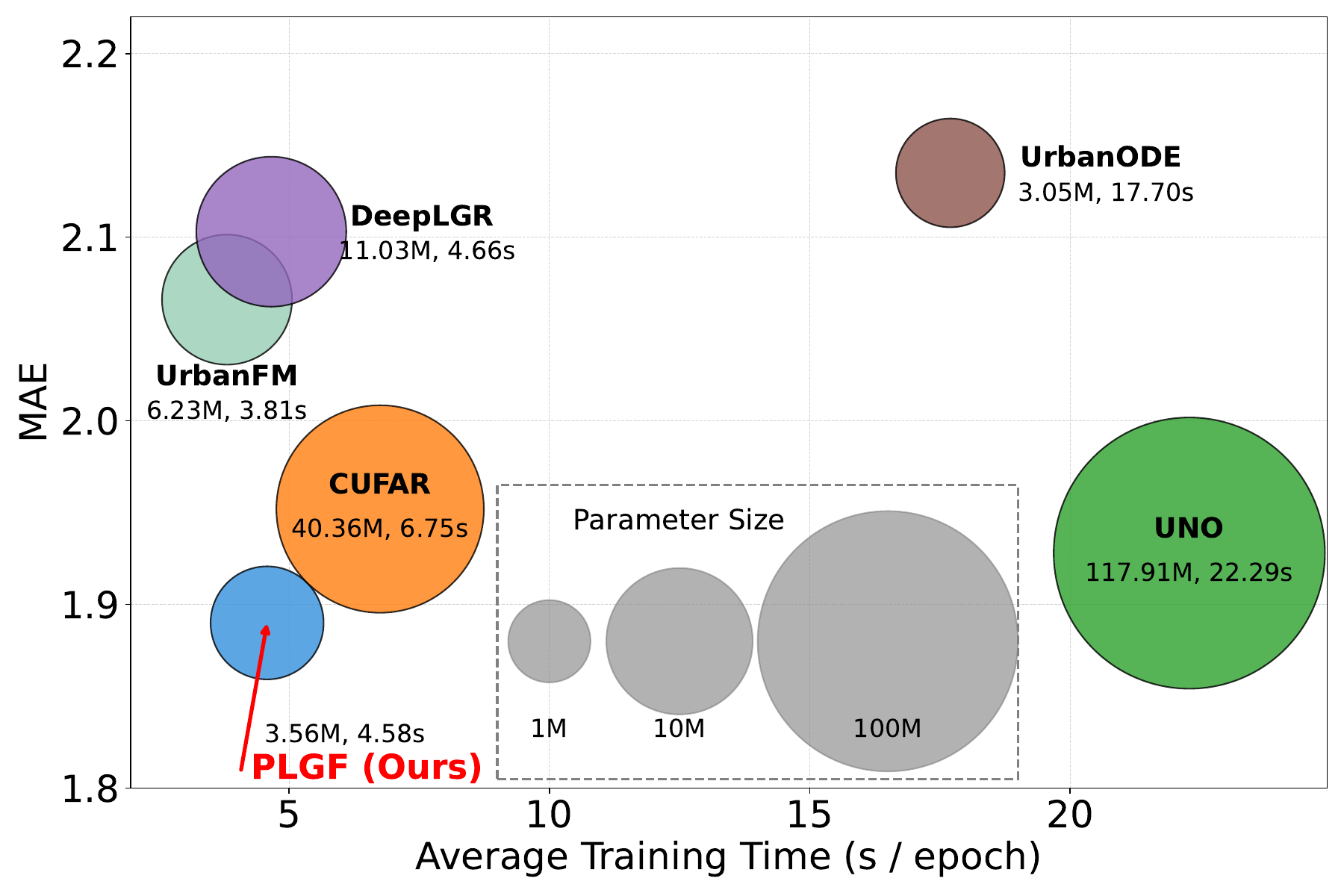} %
    \caption{Model efficiency comparison.}
    \label{fig:efficiency}
    \vspace{-3mm}
\end{figure}

Figure~\ref{fig:efficiency} provides a comprehensive efficiency comparison among representative FUFI models, where the horizontal axis indicates the average training time per epoch, the vertical axis reflects MAE performance, and the bubble size denotes the number of model parameters. As shown, our proposed PLGF achieves state-of-the-art accuracy (lowest MAE) while maintaining remarkable efficiency. Specifically, PLGF requires only 3.56M parameters and attains a 10\% lower MAE than UrbanFM (2.06) and  DeepLGR (2.10) under a comparable parameter scale, while requiring a similar or shorter training time per epoch. 
Furthermore, when compared to UNO, which achieves a similar MAE but relies on a model that is over 33 times larger (117.91M parameters) and requires nearly 5 times longer training per epoch (22.29s vs. 4.58s), PLGF demonstrates an outstanding improvement in both computational and memory efficiency.
This result highlights the superior efficiency of PLGF: it delivers advanced performance with a dramatic reduction in computational cost and model size (approximately 97\%).

\subsection{Generalizability Analysis of DualFocal Loss}
\begin{figure}[t]
    \centering
    \includegraphics[width=0.98\linewidth]{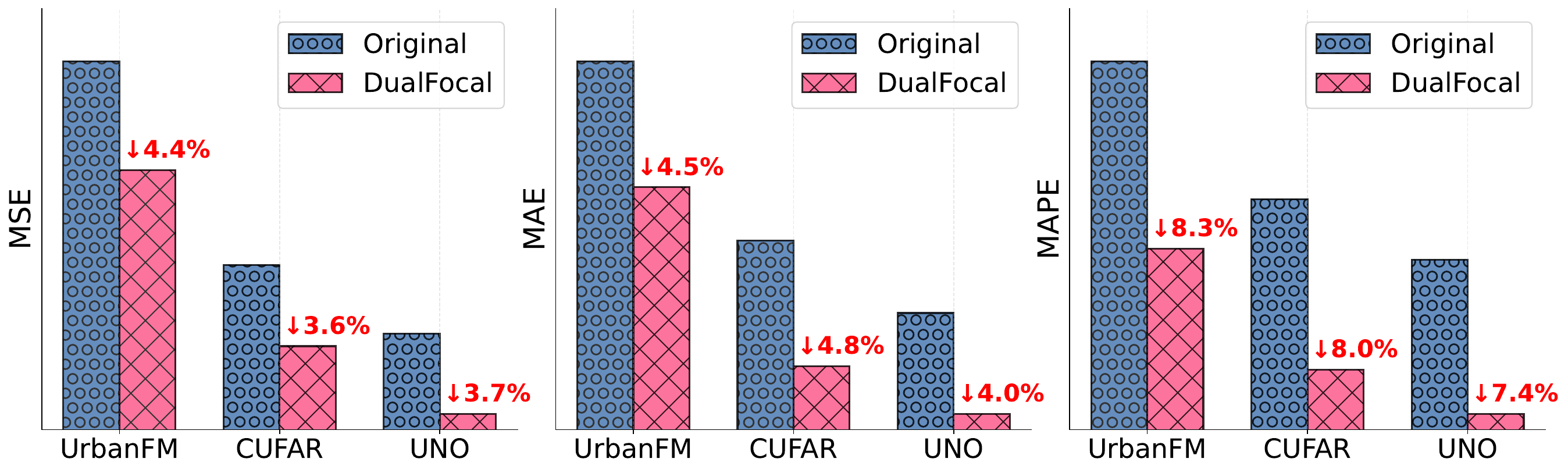} %
    \caption{Plug-and-play effectiveness of DualFocal loss. }
    \label{fig:focalloss}
\end{figure}

To further demonstrate the universality and effectiveness of our DualFocal loss, we conduct plug-and-play experiments in Task 2 by integrating it into several representative FUFI models, including UrbanFM, CUFAR, and UNO. 
As illustrated in Figure~\ref{fig:focalloss}, DualFocal consistently delivers substantial performance improvements across all evaluation metrics. Specifically, DualFocal reduces MSE by 4.4\%, 3.6\%, and 3.7\% on UrbanFM, CUFAR, and UNO, respectively, while also lowering MAE by up to 4.8\%.
Most notably, the gains in MAPE are particularly pronounced: DualFocal achieves relative reductions of 8.3\% (UrbanFM), 8.0\% (CUFAR), and 7.4\% (UNO). 
The result underscores the core design of our loss function; by focusing the optimization on difficult, high-variance samples, it successfully addresses the challenge of skewed data distribution and forces the models to pay closer attention to relative errors, which is particularly critical in low-flow regions. 
In addition, it validates DualFocal loss as a general, effective, and easily adoptable optimization strategy that can seamlessly enhance a wide range of existing models, making it a promising choice for future FUFI research and applications.

\subsection{Paramerters Studies}

\begin{figure}[t]
\centering
    \subfigure{
        \includegraphics[width=0.48\linewidth]{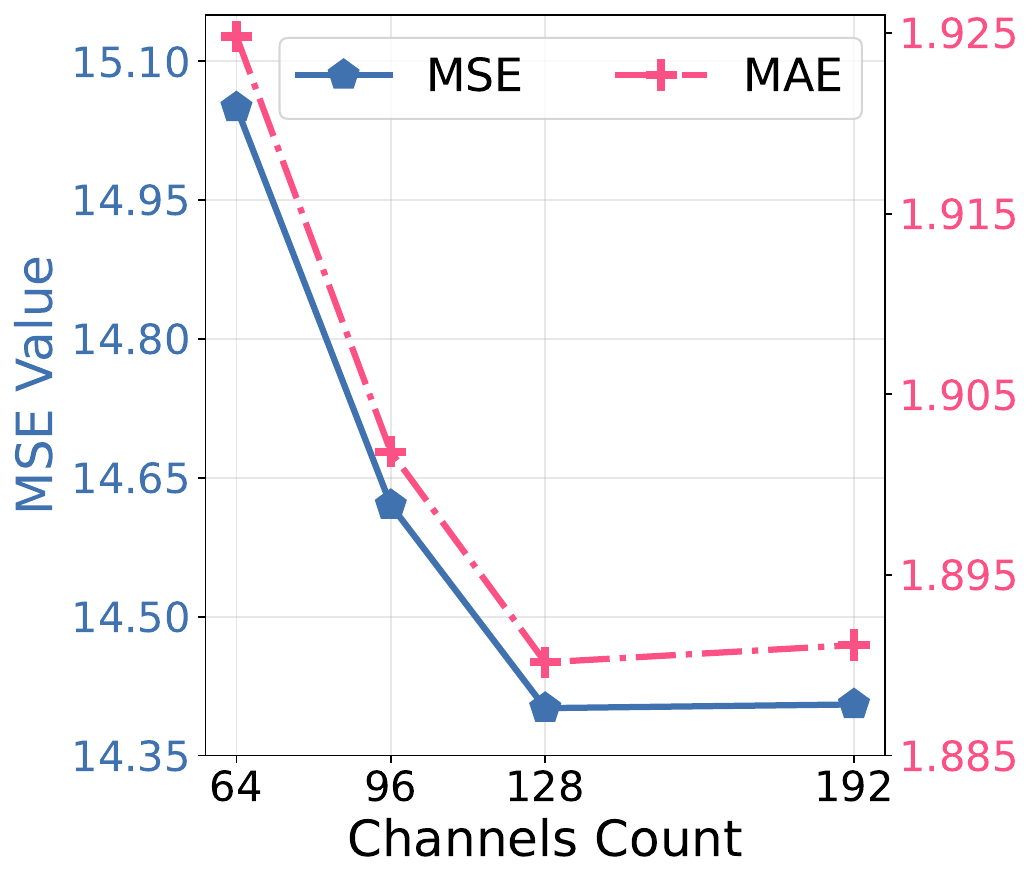}
         \label{fig:channels}
    }\hspace{-0.02\linewidth} 
    \subfigure{
        \includegraphics[width=0.48\linewidth]{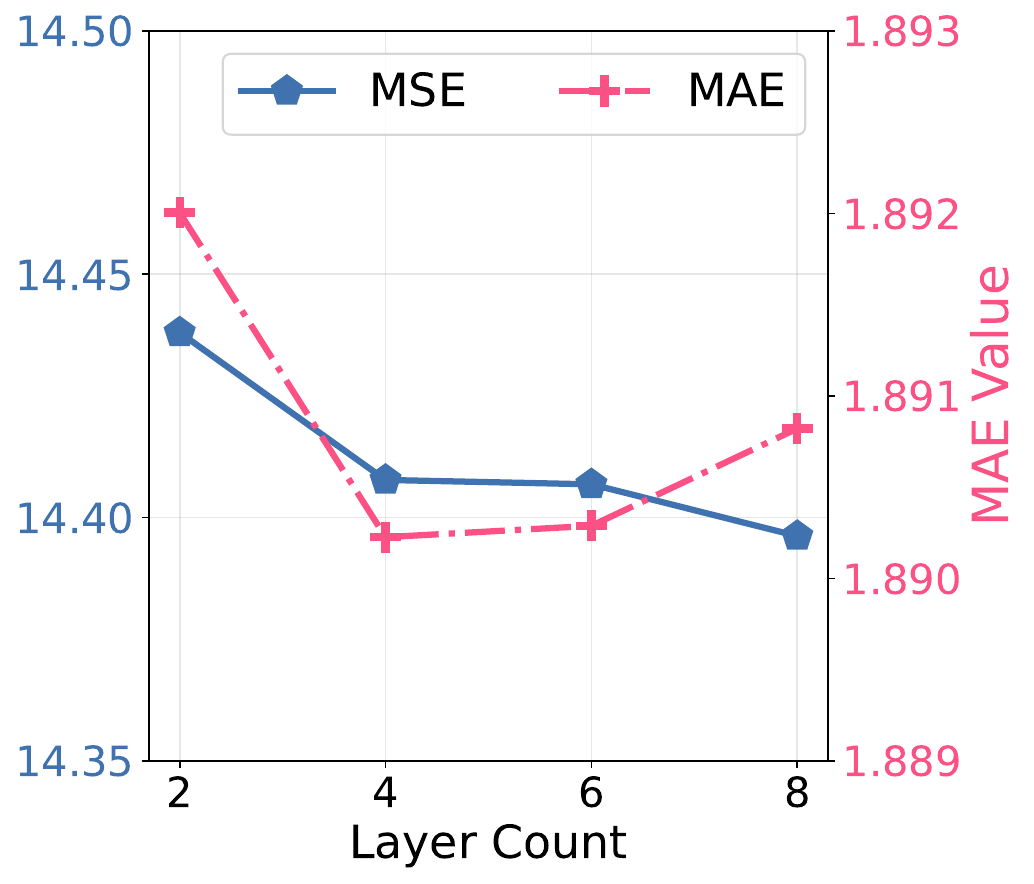}
        \label{fig:layers}
    }  
    \vspace{-5mm}
   \caption{Parameters setting and analysis.}
   \label{fig:para_analysis}
    \vspace{-3mm}
\end{figure}

To identify the optimal balance between performance and complexity, we analyzed the impact of model width (number of channels) and depth (number of local-global feature extraction layers). 
The results, summarized in Figure~\ref{fig:para_analysis}, reveal a clear trade-off for both dimensions.
In the left panel, we observe that increasing the channel count from 64 to 128 leads to a substantial reduction in both MSE and MAE, indicating that a richer feature representation significantly enhances model accuracy. However, further increasing the channel number to 192 yields only marginal gains, suggesting diminishing returns beyond a certain embedding size. 
The right panel examines the effect of altering the number of feature extraction layers. Both MSE and MAE decrease as the layer count increases from 2 to 4, reflecting the benefit of deeper hierarchical feature extraction. However, as the depth continues to grow, the improvements plateau and may even slightly fluctuate, implying that excessive depth does not necessarily translate into better performance and may introduce unnecessary computational overhead.
Overall, these results highlight the importance of carefully balancing width (e.g., 128) and depth (e.g., 4): sufficient channel capacity and moderate network depth are both crucial for achieving optimal performance without  excessive complexity.

\subsection{Ablation Studies}

To systematically evaluate the contribution of each component within PLGF, we conduct ablation experiments by incrementally removing key modules and reporting the resulting performance (Table~\ref{tab:ablation}). The findings provide clear empirical support for the core design motivations of our framework.
First, the Local-Global Feature Fusion and Context-Gated Attention modules are both crucial for accurate urban flow inference, as evidenced by the significant performance drop when either is removed. Local-Global Feature Fusion enables the model to jointly capture fine-grained spatial details and broad contextual patterns, while Context-Gated Attention ensures adaptive recalibration based on dynamic external and contextual signals. Together, they validate our motivation that effective urban flow modeling must jointly leverage multi-scale spatial information and dynamically adapt to contextual variations.
The DualFocal Loss is also essential, notably improving MAPE and effectively addressing the imbalance and long-tail distribution of urban flows, thus ensuring accurate predictions for both high-flow and low-flow regions.
Other components, such as FiLM-based feature modulation and logarithmic transformation, offer additional but smaller improvements. FiLM promotes flexible feature adaptation, while the logarithmic transformation enhances robustness to outliers, as reflected by slightly higher MSE and consistent gains in MAE and MAPE.

\begin{table}[t]
\centering
\small
\begin{tabular}{lccc}
\toprule
Method & MSE & MAE & MAPE   \\
\midrule
PLGF    & 14.408  &  \textbf{1.890} & \textbf{0.281}  \\
\cmidrule(lr){1-4}
wo/ FiLM           & 14.481  & 1.894  & 0.282  \\
wo/ Local-Global Fusion   & 17.745  & 2.049  & 0.288  \\
wo/ Context-Gated Attention  & 18.638  & 2.095  & 0.294  \\
wo/ DualFocal      & 14.574  & 1.926  & 0.304  \\
wo/ Logarithmic    & \textbf{14.349}  & 1.902  & 0.301  \\
\bottomrule
\end{tabular}
\caption{Ablation study for core component in PLGF.}
\label{tab:ablation}
    \vspace{-3mm}
\end{table}

In summary, these ablation results strongly confirm the necessity and effectiveness of our architectural choices and optimization strategies, each of which is closely rooted in the unique challenges and dynamics of the fine-grained urban flow inference task.

\section{Conclusion}

In this work, we proposed a novel lightweight framework for fine-grained urban flow inference that systematically tackles two critical barriers to practical deployment: model bloat and unfocused optimization. 
Architecturally, our PLGF model introduces a Progressive Local-Global Fusion strategy to efficiently capture complex spatial dependencies without prohibitive computational cost. 
For optimization, the proposed DualFocal Loss provides a powerful, adaptive learning mechanism that is keenly aware of the skewed data distributions and prediction difficulties inherent in real-world urban environments.

\bibliography{aaai2026}

\clearpage
\appendix

\section{Experimental Setup}
To support the reproducibility of the results presented in this study, we provides additional details regarding the experimental setup. 
We elaborate on the specifics of the datasets, the baseline models used for comparison, the formulation of our evaluation metrics, and the key implementation details of our proposed framework.

\subsection{Datasets}
Our experiments are conducted on the widely-used public TaxiBJ benchmark dataset to ensure a comprehensive and fair evaluation against prior work. 
This large-scale, real-world dataset consists of taxi trajectories collected in Beijing over four distinct time periods spanning from 2013 to 2016, which are often treated as four individual tasks (Task 1-4) in the literature.

\begin{table}[h]
\centering
\begin{tabular}{lllc}
\toprule
\textbf{Dataset}  & \textbf{Task} &  \textbf{Time Span} & \textbf{Size}   \\
\midrule
\multirow{4}{*}{TaxiBJ} & Task-1 & 7/1/2013 --- 10/31/2013  & 3,060  \\
& Task-2 & 2/1/2014 --- 6/30/2014 & 3,559  \\
& Task-3 & 3/1/2015 --- 6/30/2015 &  3,792  \\
& Task-4 & 11/1/2015 --- 3/31/2016 & 4,244  \\
\bottomrule
\end{tabular}
\caption{TaxiBJ Dataset Statistics.}
\label{tab:dataset_stats}
\end{table}

\begin{table}[h]
\centering
\begin{tabular}{llr}
\toprule
\textbf{Factor}  &  &   \textbf{Value}   \\
\midrule
Weather &  &  16 types \\
Temperature &  & [-24.6, 41.0] $^{\circ} $C  \\
Wind Speed &  &  [0, 48.6] mph  \\
Day of Week & & 7 types  \\
Time  of Day &  & 24 types \\
Holiday &  & 2 types \\
Weekend &  & 2 types \\
\bottomrule
\end{tabular}
\caption{External factors.}
\label{tab:dataset_factors}
\end{table}

For the primary FUFI task, the data is spatially organized into a coarse-grained grid of $32\times32$ and a corresponding fine-grained ground truth of $128\times128$, representing a $4\times$ upscaling factor. 
Flow volumes are aggregated into 30-minute intervals. The dataset also includes a rich set of external factors, such as weather conditions, temperature, wind speed, and temporal indicators (e.g., day of week, holidays), which are used as conditional inputs for our model. Following standard protocols, for each of the four tasks, we partition the data chronologically into training, validation, and test sets.
Table \ref{tab:dataset_stats} and Table \ref{tab:dataset_factors} detail the time span, dataset size, and external factors involved in each task.

\subsection{Baselines}

Our proposed framework is access against a wide spectrum of representative and state-of-the-art models. These baselines are divided into two main categories: general-purpose super-resolution models and FUFI-specific models.

\begin{itemize}
    \item SRCNN \cite{dong2015image}  A pioneering super-resolution model that first successfully introduced a three-layer convolutional neural network (CNN) for the SR task.
    \item ESPCN \cite{shi2016pixel} An efficient super-resolution method that employs a sub-pixel convolutional layer to perform upsampling in a single, computationally efficient step.
    \item VDSR \cite{kim2016accurate} A model that significantly improves super-resolution accuracy by utilizing a very deep convolutional network architecture with up to 20 layers.
    \item DeepSD \cite{vandal2017deepsd} A super-resolution method that adopts multi-scale inputs, originally designed for statistical downscaling of climate data.
    \item SRResNet \cite{ledig2017photo} A powerful super-resolution model that enhances deep architectures by incorporating residual learning to facilitate the training of deeper networks.
    \item UrbanFM \cite{liang2019urbanfm} The pioneering work that first formulated the FUFI problem, introducing a distributional upsampling module to handle the spatial constraint and a dedicated subnet for fusing external factors.
    \item DeepLGR \cite{liang2021revisiting} A FUFI model that refines spatial modeling by using a dual-path architecture to learn both global spatial dependencies and local feature representations simultaneously.
    \item FODE \cite{zhou2020enhancing} A FUFI model that first introduced neural ODE to capture spatial correlations from a dynamic systems perspective, aiming to overcome numerically unstable gradient computations.
    \item UrbanODE \cite{zhou2021inferring} An advanced FUFI model that builds upon the neural ODE framework, often incorporating additional architectural components such as pyramid attention networks to further enhance the modeling of urban flow dynamics.
    \item UrbanPy \cite{ouyang2020fine} An extension of UrbanFM designed for large-scale inference, which uses a cascading pyramid architecture and a multi-stage refinement pipeline.
    \item CUFAR \cite{yu2023overcoming} The first FUFI model to address the catastrophic forgetting problem in dynamic settings, using an adaptive knowledge replay strategy for continual learning.
    \item UNO \cite{gao2024enhancing} A recent FUFI model that introduces neural operator learning to create grained-invariant solutions. It also addresses catastrophic forgetting and privacy concerns through a data-free incremental learning approach.
\end{itemize}

\subsection{Metrics}
Following standard practice in the FUFI field, we adopt three common metrics to assess model performance from different facets. 
Let $\hat{X}^{f}$  be the inferred fine-grained map and ${X}^{f}$ be the ground truth map, with a total of $M$ cells (pixels). For all metrics, lower values indicate better performance.
\begin{itemize}[leftmargin=*]
    \item \textbf{Mean Squared Error (MSE):} This metric measures the average of the squares of the errors. By squaring the differences, it penalizes larger errors more heavily than MAE. It is defined as:
    \begin{equation}
    \text{MSE} = \frac{1}{M} \sum_{i,j} (\hat{x}^f_{i,j} - x^f_{i,j})^2
    \end{equation}

    \item \textbf{Mean Absolute Error (MAE):} This metric measures the average magnitude of the errors and is less sensitive to outliers than MSE. It is defined as:
    \begin{equation}
    \text{MAE} = \frac{1}{M} \sum_{i,j} |\hat{x}^f_{i,j} - x^f_{i,j}|
    \end{equation}

    \item \textbf{Mean Absolute Percentage Error (MAPE):} This metric evaluates the relative error, providing a scale-independent measure of accuracy that is particularly useful for skewed data distributions where many flow values are close to zero. It is defined as:
    \begin{equation}
    \text{MAPE} = \frac{1}{M} \sum_{i,j} \left| \frac{x^f_{i,j} - \hat{x}^f_{i,j}}{x^f_{i,j} + \epsilon} \right| \times 100\%
    \end{equation}
\end{itemize}

\subsection{Implementation}
Our proposed PLGF framework is implemented using PyTorch 3.9 and all experiments are conducted on a server with NVIDIA RTX 4090 GPUs.

\noindent \textbf{Training Details:} For model optimization, we use the Adam optimizer with an initial learning rate of $3\times 10^{-4}$. 
A learning rate decay schedule is employed, where the rate is halved every 8 epochs to ensure stable convergence. 
We train the models for a total of 100 epochs with a batch size of 20. To prevent overfitting, we apply early stopping based on the model's performance on the validation set, saving the checkpoint with the best validation score.

\noindent \textbf{Architectural Hyperparameters:} For the PLGF architecture, the base number of channels (base channels) is set to 128. 
The Residual Dense Blocks (RDB) in the local path are configured with a growth rate of 32 and num layers set to 4. The residual scaling factor $\alpha$ is initialized to 0.1. 
The EnhancedFPN in the global path is configured with num scales set to 4. For our proposed DualFocal Loss, the balancing hyperparameter $\lambda$ for the logarithmic component is set to 10, and the focalizing parameters $\beta$ and $\gamma$ are set to 0.2 and 1.0.

\section{DualFocal Loss}

\subsection{Motivation and Intuition}
Conventional regression losses are often suboptimal for the FUFI task due to the highly skewed, non-uniform distribution of urban flow data. 
The data is typically characterized by a long-tail distribution: a vast majority of grid cells have very low or zero flow (``valleys''), while a small number of cells have extremely high flow values (``peaks''). 
Our DualFocal Loss is designed to address this imbalance with: (1)\textbf{ Scale-Aware Supervision} to be sensitive to errors in both peaks and valleys, and (2) \textbf{Hard Example Focusing} to concentrate on the most informative samples.

\subsection{Formal Definition}
The DualFocal Loss is formulated as a modulating factor applied to a dual-scale loss. Given a prediction  $\hat{X}^f$ and a ground truth $X^f$, the loss is defined as:
\begin{equation}
\mathcal{L}_{\text{Dual-Focal}} = (f(\beta \cdot \mathcal{L}_{\text{ds}}))^\gamma \cdot \mathcal{L}_{\text{ds}}
\end{equation}
where the loss $\mathcal{L}_{\text{ds}}$ is composed of two parts:
\begin{equation}
\mathcal{L}_{\text{ds}} = \underbrace{|\hat{X}^f - X^f|}_{\mathcal{L}_{\text{l1}}} + \lambda \cdot \underbrace{|\log(1+\hat{X}^f) - \log(1+X^f)|}_{\mathcal{L}_{\text{log}}}
\end{equation}

\subsection{Justification of Dual-Scale Supervision}
The loss $\mathcal{L}_{\text{ds}}$ is a combination of a standard L1 loss ($\mathcal{L}_{\text{l1}}$) and a logarithmic L1 loss ($\mathcal{L}_{\text{log}}$). An analysis of their gradients with respect to a prediction $\hat{y}$ for a true value $y$ reveals their complementary nature.
\begin{itemize}[leftmargin=*]
    \item \textbf{Linear Scale:} The loss is $\mathcal{L}_{\text{l1}} = |y - \hat{y}|$, and gradient is:
    \begin{equation}
    \frac{\partial \mathcal{L}_{\text{l1}}}{\partial \hat{y}} = -\text{sgn}(y - \hat{y}).
    \end{equation}
    The magnitude of this gradient is always 1. While effective for penalizing large absolute errors, it is insensitive to the large relative errors that occur in low-flow regions.

    \item \textbf{Logarithmic Scale:} The loss is $\mathcal{L}_{\text{log}}=|\log(1+y) - \log(1+\hat{y})|$. Its gradient is:
    \begin{equation}
    \frac{\partial \mathcal{L}_{\text{log}}}{\partial \hat{y}} = - \frac{1}{1+\hat{y}} \cdot \text{sgn}(\log(1+y) - \log(1+\hat{y})).
    \end{equation}
    The key term here is  $\frac{1}{1+\hat{y}}$. The gradient magnitude is now inversely proportional to the predicted value, making the model highly sensitive to prediction errors in low-flow regions (small $\hat{y}$). 
\end{itemize}
By combining both, $\mathcal{L}_{\text{ds}}$ receives a balanced gradient signal that supervises both absolute and relative error, making it inherently scale-aware.

\subsection{Focalizing Mechanism}
The complete DualFocal Loss applies a dynamic modulating factor to the dual-scale loss, $M(\mathcal{L}_{\text{ds}}) =  (f(\beta \cdot \mathcal{L}_{\text{ds}}))^\gamma$, where $f(\cdot)$ is a modulating function (e.g., tanh functaion).
For easy samples where the loss $\mathcal{L}_{\text{ds}} \rightarrow 0$, the modulating factor $M \rightarrow0$, significantly down-weighting the sample's loss. For hard samples, $M \rightarrow 1$, and the total loss remains high. 
This mechanism's primary effect is on the gradient flow; for easy samples, the gradient contribution is suppressed, while for hard samples, the gradient is allowed to pass through with its full magnitude. In summary, the focalizing mechanism mathematically achieves a form of hard example mining, forcing the model to focus its learning capacity on the samples it finds most difficult.

The DualFocal Loss has three key hyperparameters. $\lambda$: Balances the importance of absolute error ($\mathcal{L}_{\text{l1}}$) versus relative error ($\mathcal{L}_{\text{log}}$).
$\beta$:  Controls how quickly the focal mechanism activates.
$\gamma$: Controls the aggressiveness of the focusing.
In our experiments, these were set to $\lambda=10, \beta=0.2, \gamma=1.0$ based on empirical validation, which we found to provide a stable and effective optimization process.

\begin{figure*}[h]
    \centering
    \includegraphics[width=0.96\linewidth]{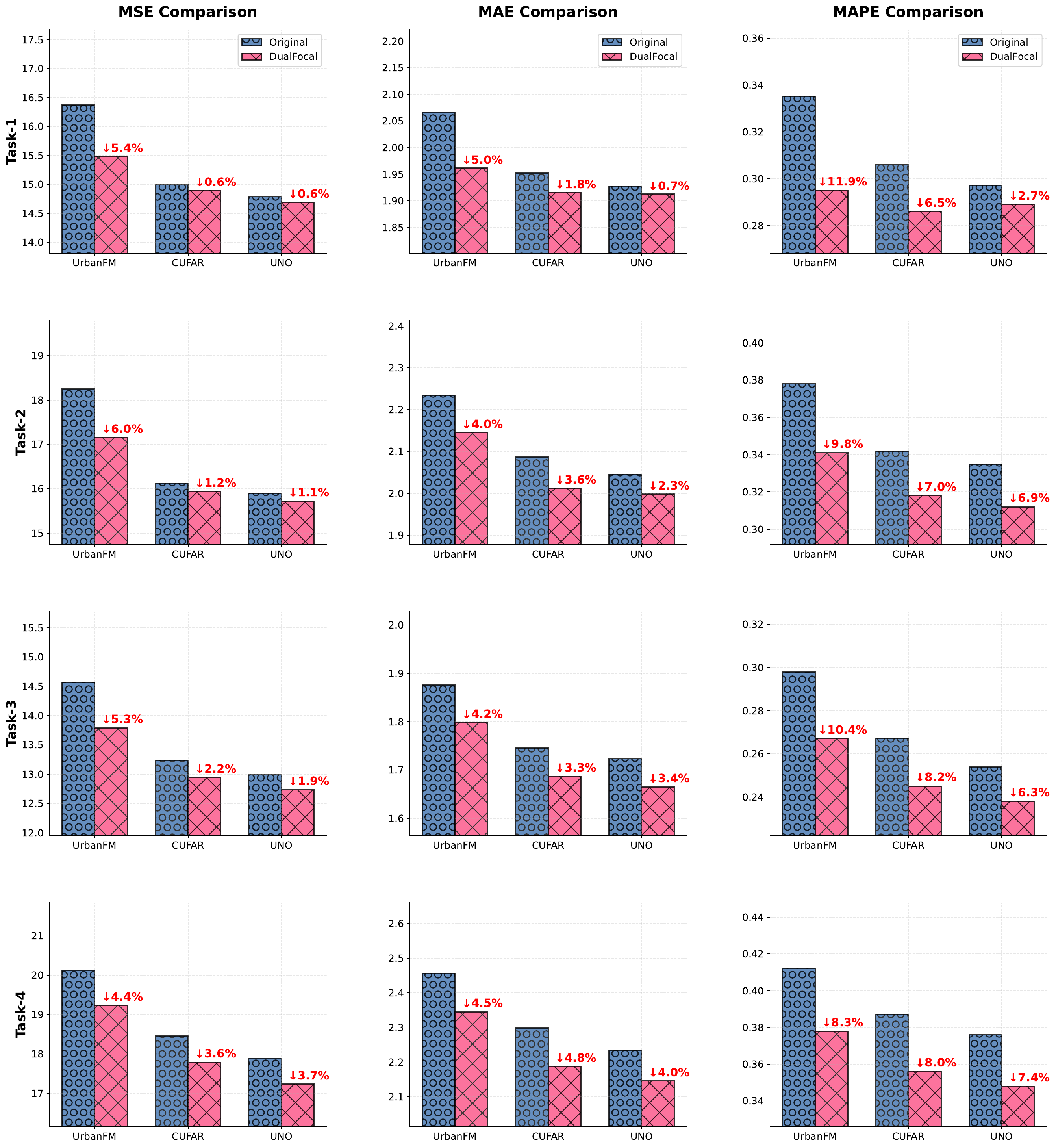} %
    \caption{Plug-and-play effectiveness of DualFocal loss for all Tasks.}
    \label{fig:suppl_exps}
\end{figure*}

\end{document}